\newif\ifgcpr
\gcprfalse

\newif\ifreview
\reviewtrue

\ifgcpr

\documentclass[runningheads]{llncs}

\usepackage{graphicx}
\usepackage{comment}
\usepackage{amsmath,amssymb}
\usepackage{color}
\usepackage{url}
\usepackage{hyperref}

\usepackage[utf8]{inputenc} 
\usepackage[T1]{fontenc}    

\ifreview
	\usepackage{lineno}

	\linenumbers
\fi

\else
\documentclass[onecolumn]{IEEEtran} 

\usepackage{times}
\usepackage[utf8]{inputenc} 
\usepackage[T1]{fontenc}    
\usepackage{hyperref}       
\usepackage{url}            
\usepackage{booktabs}       
\usepackage{amsfonts}       
\usepackage{nicefrac}       
\usepackage{microtype}      

\usepackage[numbers]{natbib}
\fi

\usepackage[disable]{todonotes}
\usepackage{comment}
\usepackage{tabularx}
\usepackage{wrapfig}
\usepackage[capitalise]{cleveref}

\newcommand{\todominor}[2][]{\todo[color=yellow, #1]{\noindent\footnotesize #2}}

\newcommand{\capenumfont}[1]{{\bf #1)}}

\newcommand{\camvid}{\texttt{CamVid}}
\newcommand{\cityscapes}{\texttt{Cityscapes}}

\newcommand{\numlayers}{N}
\newcommand{\numlayersindex}{n}
\newcommand{\numblocks}{L}

\newcommand{\numrepeats}{K}

\newcommand{\numclasses}{C}
\newcommand{\semmaptrue}{\tilde{m}_\numlayersindex}
\newcommand{\semmap}{m_\numlayersindex}
\newcommand{\crossentropy}{c_\numlayersindex}
\newcommand{\lossweight}{a_\numlayersindex}
\newcommand{\lossweighti}{a_i}
\newcommand{\numoutchan}{f}
\newcommand{\numoutchanres}{\numoutchan_\mathrm{r}}

\begin{document}

\title{ItNet: iterative neural networks with small graphs for accurate, efficient and anytime semantic segmentation}

\ifgcpr
\else
\author{Thomas Pfeil \\
Bosch Center for Artificial Intelligence \\
Renningen, Germany \\
{\tt\small thomas.pfeil@de.bosch.com}}

\date{\today}
\fi

\maketitle

\begin{abstract}
Deep neural networks have usually to be compressed and accelerated for their usage in low-power, e.g.\ mobile, devices.
Recently, massively-parallel hardware accelerators were developed that offer high throughput and low latency at low power by utilizing in-memory computation.
However, to exploit these benefits the computational graph of a neural network has to fit into the in-computation memory of these hardware systems that is usually rather limited in size.
In this study, we introduce a class of network models that have a small memory footprint in terms of their computational graphs.
To this end, the graph is designed to contain loops by iteratively executing a single network building block.
Furthermore, the trade-off between accuracy and latency of these so-called iterative neural networks is improved by adding multiple intermediate outputs during both training and inference.
We show state-of-the-art results for semantic segmentation on the \camvid{} and \cityscapes{} datasets that are especially demanding in terms of computational resources.
In ablation studies, the improvement of network training by intermediate network outputs
as well as the trade-off between weight sharing over iterations and the network size are investigated.
\end{abstract}

\section{Introduction}\label{intro}
\label{sec:intro}

For massively-parallel hardware accelerators \cite{Schemmel2010_waferscale,Merolla2014_truenorth,Peng2020_memristor,graphcore,mythic}, every neuron and synapse in the network model has its physical counterpart on the hardware system.
Usually, by design, memory and computation is not separated anymore, but neuron activations are computed next to the memory, i.e.\ the parameters, and fully in parallel.
This is in contrast to the rather sequential data processing of CPUs and GPUs, for which the computation of a network model is tiled and the same arithmetic unit is re-used multiple times for different neurons.
Since the computation is performed fully in parallel and in memory, the throughput of massively-parallel accelerators is usually much higher than for CPUs and GPUs.
This can be attributed to the fact that the latency and power consumption for accessing local memory, like for in-memory computing, are much lower than for computations on CPUs and GPUs that require the frequent access to non-local memory like DRAM \cite{sze2017_power}.
However, the network graph has to fit into the memory of the massively-parallel hardware accelerators to allow for maximal throughput.
If the network graph exceeds the available memory, in principle, the hardware has to be re-configured at high frequency to sequentially process the partitioned graph, as it is the case for CPUs and GPUs, and the benefit in terms of high throughput would be substantially reduced or even lost.
Even higher throughputs can be obtained by using mixed-signal massively-parallel hardware systems that usually operate on shorter time scales than digital ones, e.g., compare \cite{Schemmel2010_waferscale,Peng2020_memristor,mythic} to \cite{Merolla2014_truenorth,graphcore}.

\begin{figure}[t]
  \centering
  \ifgcpr
\else
  \raisebox{0.1in}{\includegraphics[scale=1.0]{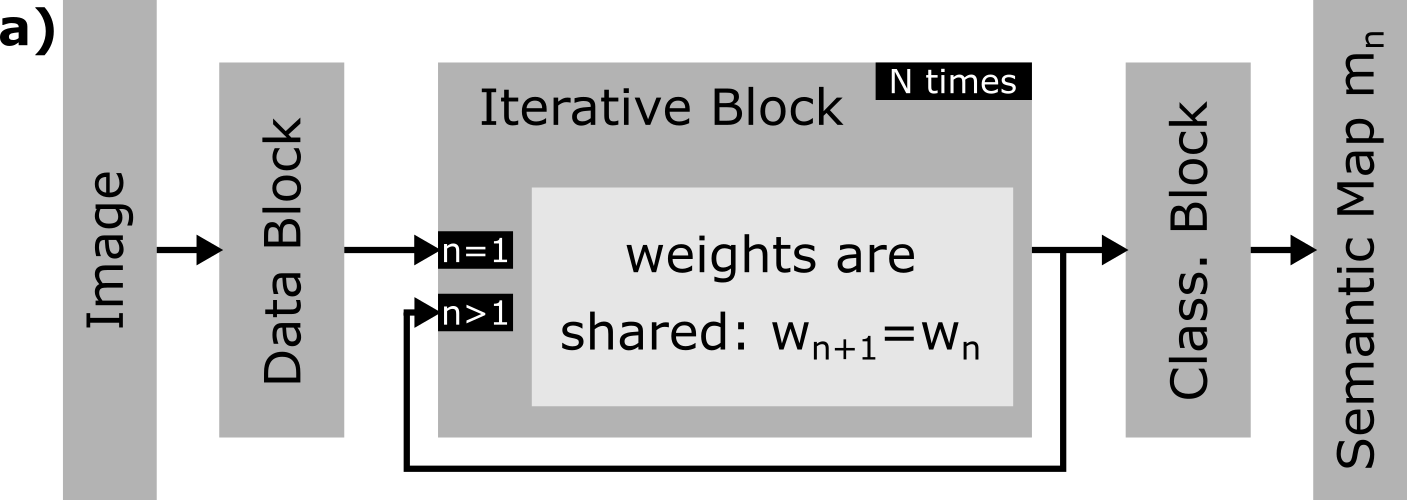}}
  \includegraphics[scale=1.0]{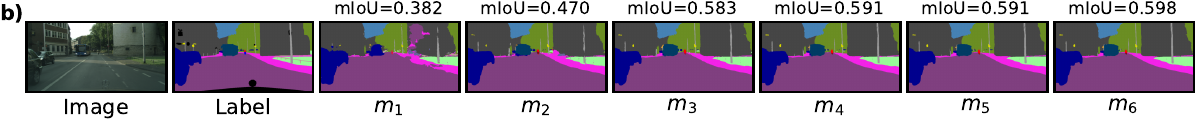}
  \fi

  \caption{
    The iterative neural network (ItNet):
    first, images are pre-processed and potentially down-scaled by a sub-network called \emph{data block}.
    Then, the output of this data block is processed by another sub-network that is iteratively executed.
    After every iteration $\numlayersindex$, the output of this \emph{iterative block} is fed back to its input and, at the same time,
    is further processed by the \emph{classification block} to predict the semantic map $\semmap$.
    This network generates multiple outputs $\semmap$ with increasing quality (for an example, see bottom) and computational costs
    and heavily re-uses intermediate network activations.
  }
  \label{fig:intro}
\end{figure}
In order to achieve neural networks with small computational graphs,
in which nodes are operations and edges are activations,
we heavily re-use a single building block of the network (see the \emph{iterative block} in \Cref{fig:intro}a).
Not only the structure of computations, i.e.\ the type of network layers including their dimensions and connectivity, is identical for each iteration of this building block,
but also the parameters are shared between iterations.
In the computational graphs of these so-called \emph{iterative neural networks} (ItNets), the re-used building blocks with shared weights can be represented by nodes with self-loops.
Compared to conventional feed-forward networks, loops simplify the graph by reducing the number of unique nodes and, consequently, its computational footprint.
However, the restriction of sharing weights usually decreases the number of free parameters and, hence, the accuracy of networks.
To isolate and quantify this effect we compare networks with weight sharing to networks, for which the parameters of the building blocks are chosen to be independent between iterations of the building block.
In contrast to the above proposal, conventional deep neural networks for image processing usually do not share weights and have no (e.g., \cite{Huang2017_densenet}) or few (e.g., one building block for each scale like by \cite{greff2016_highway}) layers of identical structure.
\cite{qianli2016_resnetrnn} share weights between re-used building blocks, but use multiple unique building blocks.

To improve the training of networks, which contain loops in their graphs, and to reduce the latency of networks during inference we use multiple intermediate outputs.
By progressively updating the network prediction over the network outputs this so-called \emph{anytime prediction} allows to trade off accuracy versus latency in an online manner with barely any overhead (e.g., \cite{amthor2016,Huang2017_multiscaledensenet}).

If all nodes in the network graph are computed in parallel, like on massively-parallel hardware systems,
the latency for inference is dominated by the depth of the network, i.e. the longest path from input to output \cite{fischer2018_streaming}.
In order to reduce this latency and to allow the repetitive execution of a single building block,
we use networks that compute all scales in parallel (similar to \cite{Huang2017_multiscaledensenet,Ke2017_multigrid}) and increase the effective depth for each scale by consecutive iterations of this multi-scale building block.
Furthermore, multi-scale networks are also beneficial for the integration of global information, as especially required by dense prediction tasks like semantic segmentation \cite{Zhao2017_pspnet}.
To further reduce the latency we do not only compute all scales in parallel, but also keep the depth of each scale as shallow as possible.
In other words, we choose the number of layers in the longest path from input to output of the consecutively executed building block as small as possible.

In deep learning literature, the computational costs are usually quantified by counting the parameters and/or the multiply-accumulate operations (MACs) required for the inference of a single sample.
For fully convolutional networks, the number of parameters is independent of the spatial resolution of the network's input and the intermediate feature maps.
Especially for large inputs as commonly used for semantic segmentation,
the number of parameters does not cover the main workload and is, hence, not suited as a measure for computational costs.
MACs have the advantage that they can be easily calculated and are usually a good approximation for the latency and throughput on CPUs and even GPUs.
However, for most novel hardware accelerators, not the MACs, but the non-local memory transfers are dominating the computational costs in terms of power consumption \cite{Chen2016_eyeriss,sze2017_power,Chao2019_hardnet}.
These memory transfers are minimized on massively-parallel hardware systems as long as the network graph fits into the \emph{in-computation memory} of these systems, i.e.\ the memory of their arithmetic units.
Since both the power consumption during inference and the production cost scale with the size of this memory we additionally compare the size of the computational graphs between networks.
Note that the practical benefits of ItNets cannot be demonstrated on conventional CPUs or GPUs, since these hardware systems do not support the processing of neural networks in a
fully-parallel and, hence, low-latency fashion.

This study focuses on network models in the low-power regime of only few billion MACs and addresses the resource-intensive scenario of semantic segmentation of large-scale images (for datasets, see \Cref{sec:data}).
ENet \cite{paszke2016_enet}, ESPNetv2 \cite{Mehta2019_espnetv2}, CGNet \cite{wu2018_cgnet}, and RecUNet \cite{Wang_2019_ICCV} denote efficient reference networks for this low-power regime and these datasets.
ENet and ESPNetv2 are classical convolutional neural networks, whereas the latter, to our knowledge, is the state-of-the-art in terms of mIoU over MACs.
RecUNet shares the weights of U-Nets and recurrently connect their bottlenecks.
CGNet integrates context by using dilated convolutions and global context extractors, which could also improve the performance of ItNets.

In contrast to conventional architectures of deep neural networks that were developed and optimized for their execution on rather sequential GPUs, in this study:
\begin{itemize}
  \item We present an approach to design efficient network architectures with small computational graphs that potentially allows to exploit the benefits of high throughput, low latency and low power offered by novel massively- or even fully-parallel hardware systems (\Cref{sec:netarch,sec:discussion}).
  \item We investigate in isolation the trade-off between the network’s accuracy and the re-usage of a single building block within this network (\Cref{sec:eval,sec:nas,sec:kpis}).
  \item We show the benefit of using multiple network outputs during network training and inference (\Cref{sec:training,sec:ablation}).
  \item To our knowledge, we set the new state-of-the-art in terms of accuracy over the size of the computational graph (\Cref{sec:kpis}).
\end{itemize}

We will release the source code upon acceptance for publication.

\section{Methods}\label{methods}
The following networks process images of size $x \times y \times 3$ and output $\numlayers$ semantic maps $\semmap$ of size $x \times y \times \numclasses$ with $\numclasses$ being the number of classes.

\subsection{Network architecture}
\label{sec:netarch}
\begin{figure}[t]
  \centering
  \ifgcpr
\else
  \includegraphics[scale=1.0]{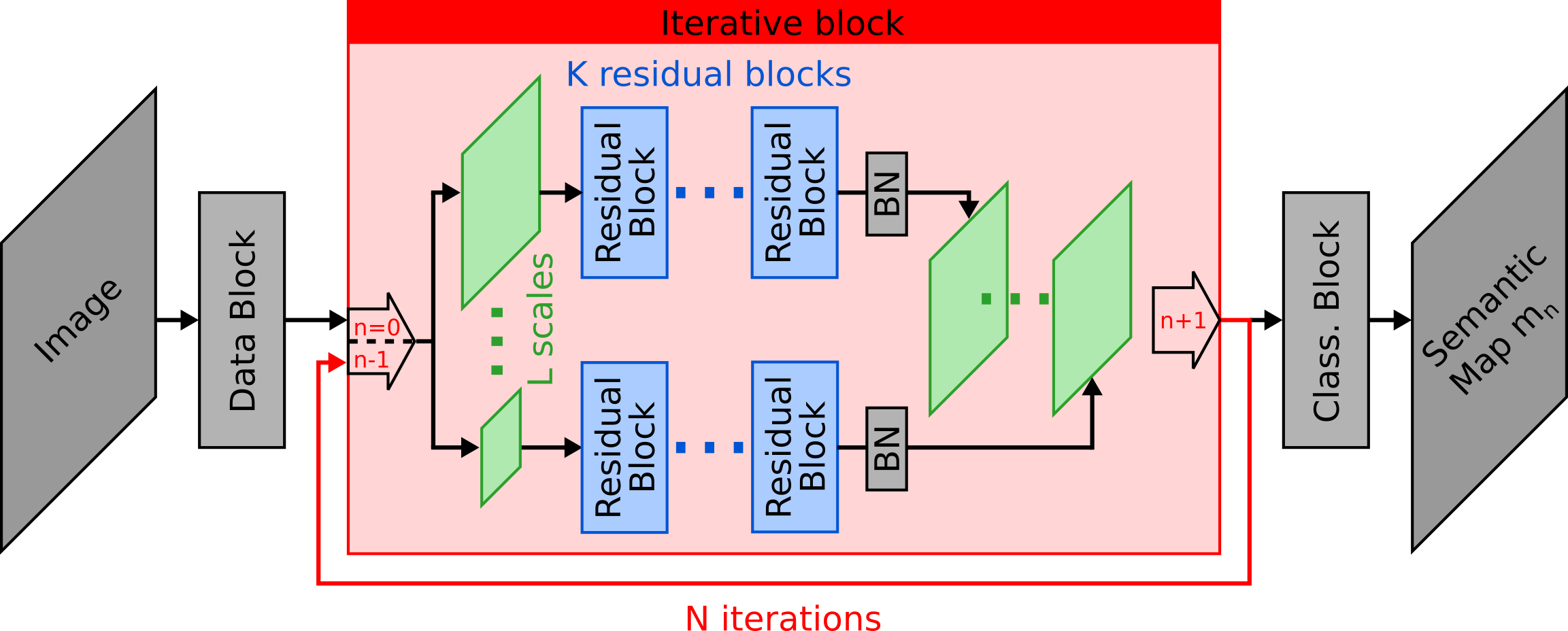}
  \fi
  \caption{
    Architectural hyperparameters of ItNets:
    The input of the iterative block is processed in parallel on $\numblocks$ different spatial scales.
    For each scale, feature maps are sequentially processed by $\numrepeats$ bottleneck residual blocks \cite{Sandler2018_mobilenetv2}.
    For a total number of $\numlayers$ iterations, the output of the iterative block is used to predict semantic maps $\semmap$ and is fed back to the next iteration $\numlayersindex$ of the iterative block.
    All parameters of the iterative block are shared between iterations $\numlayersindex$, except these of the  batch-normalization layers (BN; similar to \cite{Guo2019_CVPR}).
  }
  \label{fig:methods}
\end{figure}

We introduce a class of neural networks with small computational graphs by heavily re-using intermediate activations and weights (\Cref{fig:intro}a).
Conceptionally, the network model can be split into three main building blocks: the \emph{data block}, the \emph{iterative block} and the \emph{classification block} (for details, see \Cref{fig:supp_methods}).
While the data block is executed only once for each image, the iterative block can be executed multiple times in a row by feeding back its output as the input for the next iteration.
The classification block outputs the prediction of the semantic map by processing the intermediate activations of the feedback signal.
While the weights of the iterative block are shared between iterations, the weights of the classification block are unique for each iteration.
For comparability with previous studies, we limit the number of MACs and select the network with the highest accuracy by optimizing the architectural hyperparameters as visualized in \Cref{fig:methods}:
the number of scales $\numblocks$, the number of bottleneck residual blocks $\numrepeats$, and the number of iterations $\numlayers$.

\subsection{Network training}
\label{sec:training}

For training, we use a joint cost function for all outputs of the network:
\begin{equation}
  \mathcal{L}=\sum_{\numlayersindex} \bar{a}_\numlayersindex \crossentropy(\semmaptrue, \semmap),
  \label{eq:loss}
\end{equation}
where $\crossentropy$ is the categorical cross entropy between the true labels $\semmaptrue$ and the network predictions $\semmap$.
The weight factors $\lossweight$ are normalized as follows: $\bar{a}_\numlayersindex=\lossweight/(\sum_i{\lossweighti})$.

We use the Adam optimizer with $\beta_1=0.9$, $\beta_2=0.999$ and a learning rate $0.001$ that we multiply with $0.1$ after $70\%$ and $85\%$ of the number of overall training epochs.
We use a batch size of $8$ and train the network for $2000$ epochs ($4000$ epochs for \Cref{fig:results_kpi}) and $900$ epochs for the \camvid{} and \cityscapes{} datasets, respectively.
For \Cref{fig:results_size}, \Cref{fig:results_ablation} and the appendix, we report the mean values and the errors of the means across $5$ trials.
For \Cref{fig:results_kpi}, we report the trial with the highest peak accuracy over $3$ trials.

For the results shown in \Cref{fig:results_kpi},
we use dropout with rate $0.1$ after the depth-wise convolutions in the bottleneck residual blocks
and an L2 weight decay of $10^{-5}$ in all convolutional layers.
For all other results, we do not use dropout and weight decay.

\subsection{Network evaluation}
\label{sec:eval}

Throughout this study, we measure the quality of semantic segmentation by calculating the mean intersection-over-union (mIoU \cite{jaccard1912_miou}), which is the ratio of the area of overlap and the area of union
\[
  IoU = \frac{\tilde{m} \cap m}{\tilde{m} \cup m}
\]
averaged over all classes.

We consider a network to perform well if it achieves a high mIoU while requiring few MACs.
To this end, we calculate the area under the curve of the mIoU ($y_\numlayersindex$) over MACs ($x_\numlayersindex$) with output index $\numlayersindex$ as follows:
\begin{equation}
  \overline{AUC}=\sum_{\numlayersindex=0}^{\numlayers-1}{
		(x_{\numlayersindex+1}-x_\numlayersindex)
		\left(\frac{(y_{\numlayersindex+1}-y_0)+(y_\numlayersindex-y_0)}{2}\right)
	}
  \label{eq:auc}
\end{equation}
with $(x_0, y_0) = (0, 0.00828)$, where $y_0$ denotes the mIoU at chance level for the \camvid{} dataset.
To compensate for different maximum numbers $x_N$ of MACs for different sets of hyperparameters, we normalize as follows: $AUC=\frac{\overline{AUC}}{x_N}$.
\todominor[inline]{add visualization}

The size of the computational graph is computed by accumulating the memory requirements of all nodes, i.e.\ network layers, in the network graph.
For each unique layer, the total required memory is the sum of the memory for parameters, input feature maps and output feature maps.
Repeated executions of the same layer, e.g.\ iterative building blocks with weight sharing, are not considered as unique and, hence, do not contribute to the size of the computational graph.
This is motivated by the fact that massively-parallel systems, especially analog hardware \cite{Peng2020_memristor}, are usually composed of cross-bar arrays, in which the weights are stored at the intersection of input and output lines.
Under the assumption that these weights cannot be re-configured during inference, the required size of a cross-bar array increases with the number of weights, but is independent from the number of iterations this cross-bar array is used.
If weights are not shared between iterations of the iterative block, the corresponding cross-bar arrays cannot be re-used, but larger or more cross-bar arrays are required or, in other words, the size of the computational graph increases.

The latency of a network, if executed fully in parallel, is determined by the accumulated latencies of all layers along the longest path from input to output of the network.
Since the real latency to execute a network layer highly depends on the hardware system, we express the latency of networks in terms of the number of executed layers, i.e. the depth of the network (see also \Cref{sec:intro}).
In case of fully-parallel hardware systems, the real latency is comparable across layers and, hence, the depth of the network is proportional to the real latency.

For both the size of the computational graph and the latency, we only consider convolutional layers like commonly done in literature (e.g., \cite{paszke2016_enet,wu2018_cgnet,Mehta2019_espnetv2}).
This means, we ignore other network layers like normalization, activations, concatenations, additions and spatial resizing, for which we assume that they can be fused with the convolutional layers.

\subsection{Datasets}
\label{sec:data}

The \camvid{} dataset \cite{Brostow2008_camvid} consists of $701$ ($367$ for training, $101$ for validation, $233$ for testing) annotated images filmed from a moving car.
We use the same static pre-processing as in \cite{Badrinarayanan2017_segnet} to obtain images and semantic labels of size $480 \times 360$
and normalize the pixel values to the interval $[0, 1]$ by dividing all pixel values by $255$.
For online data augmentation during training, we horizontally flip the pairs of images and labels at random and use random crops of size $480 \times 352$.

The \cityscapes{} dataset \cite{Cordts2016_Cityscapes} consists of $3475$ ($2975$ for training, $500$ for validation) annotated images and we use the validation set for testing.
We resize the original images and semantic labels to $1024 \times 512$ and divide all pixel values by $127.5$ and subtract $1$ to obtain values in the interval $[-1, 1]$.
For online data augmentation during training, we horizontally flip the pairs of images and labels at random.

For both datasets, pixels with class labels not marked for training are ignored in the cost function.

\subsection{Batch normalization}
\label{sec:bn}
\begin{figure}[t]
  \centering
    \ifgcpr
\else
    \includegraphics[scale=1.0]{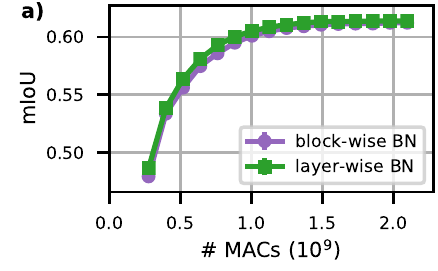}
    \includegraphics[scale=1.0]{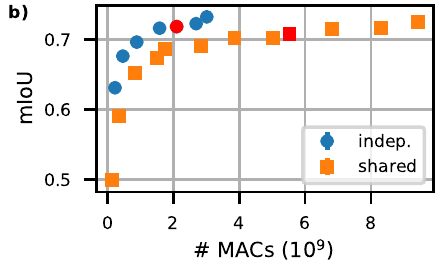}
    \fi
  \caption{
    \capenumfont{a}
      For ItNets with independent weights, mIoUs over MACs are shown for ItNets with batch normalization applied between iterative blocks (purple curve) and applied after each convolution (green curve) on the test set of the \camvid{} dataset.
    \capenumfont{b}
      For both cases of independent and shared weights in the iterative block, the peak mIoU over MACs is shown for networks with different widths on the validation set of the \camvid{} dataset.
      Each data point represents the peak mIoU over all network outputs for one specific network width and the corresponding MACs for this output.
  }
  \label{fig:results_size}
\end{figure}

In case of independent parameters between iterations of the iterative block, i.e.\ $w_{n+1}$ is independent from $w_n$ in \Cref{fig:intro}a, batch normalization improves the network training.
However, in case of weight sharing ($w_{n+1}=w_{n}$), also sharing the parameters of batch normalization between iterations significantly worsens the network training.
Since not sharing batch-normalization parameters would violate our idea to re-use the identical building block again and again, we place batch-normalization layers between the iterations of the iterative block (see \Cref{fig:methods}).
For comparability, we use the same setup also for networks without weight sharing, although the average validation mIoU is slightly decreased compared to networks that instead use batch normalization after each convolution (\Cref{fig:results_size}a).

\section{Results}\label{results}
In order to obtain accurate networks with low computational costs, we first search for the best set of the architectural hyperparameters $\numblocks$, $\numlayers$ and $\numrepeats$ (\Cref{sec:nas}).
Since we are also interested in the trade-off between weight sharing and the networks size, we also consider networks without weight sharing in this search.
Then, we investigate the impact of intermediate losses on the network performance to find the best set of weight factors for the loss function (\Cref{sec:ablation}).
Finally, for the found set of hyperparameters and weight factors, we show results of ItNets on the \camvid{} and \cityscapes{} dataset (\Cref{sec:kpis}).

\subsection{Search for architectural hyperparameters}
\label{sec:nas}

To find accurate networks with low computational costs
we perform a grid search over the hyperparameters of our network model as described in \Cref{fig:methods}.
For each set of hyperparameters, we choose the largest possible number of channels $\numoutchan$ that results in a network with less than $2.2$ billion MACs for the last output of the network on single samples of the \camvid{} dataset.
This number is motivated by the number of MACs required by state-of-the-art reference networks for \camvid{} \cite{wu2018_cgnet}.

Since we are interested in network architectures with a high mean intersection-over-union (mIoU) and a low number of MACs, we sort the architectures by their area under the curve (see \Cref{eq:auc}) for both networks with independent and shared weights.
Networks with a peak performance $\max_{n} (mIoU_\numlayersindex) \leq mIoU_{\mathrm{totmax}}-0.02$, where $mIoU_{\mathrm{totmax}}$ is the highest peak performance over all sets of hyperparameters, are considered as not suitable for applications and are discarded.
For comparability between networks with independent and shared weights, we choose the architecture with the lowest average rank across the two classes of networks and obtain the following set of hyperparameters: $\numlayers=16$ iterations, $\numblocks=3$ scales and $\numrepeats=1$ bottleneck residual blocks (for the results of the grid search, see \Cref{fig:grid_non_shared,fig:grid_shared} in the appendix).
Note that many iterations ($\numlayers=16$) result in a high re-usage of the iterative block, a low latency and a small computational graph.

\subsection{Improvement of training by using multiple outputs}
\label{sec:ablation}
\begin{figure}[t]
  \centering
  \ifgcpr
\else
    \includegraphics[scale=1.0]{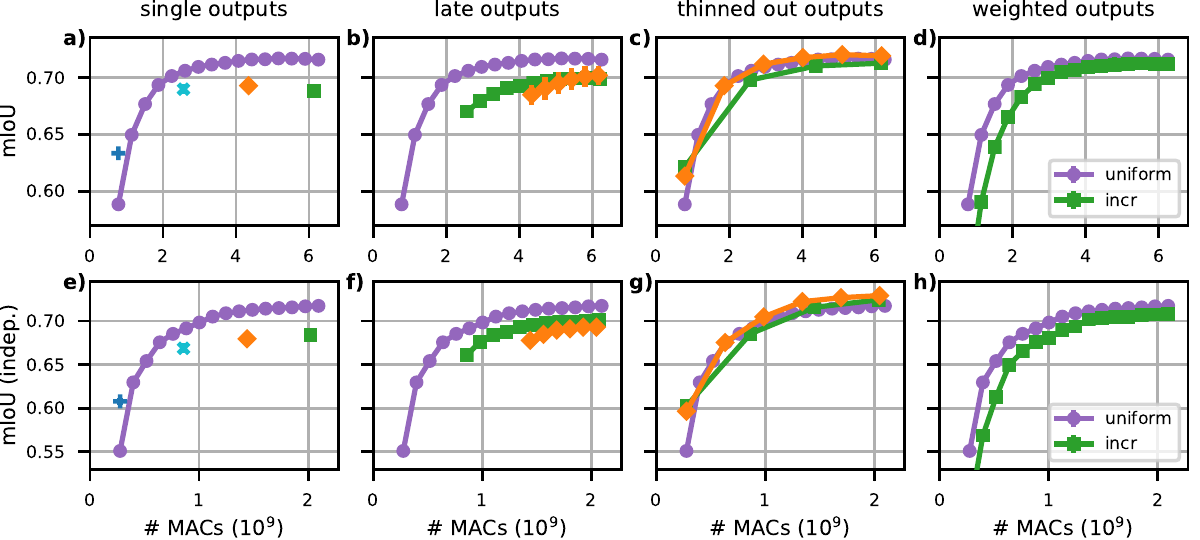}
  \fi
  \caption{
    Ablation studies for multi-output training of ItNets with (top row) and without (bottom row) weight sharing on the validation set of the \camvid{} dataset.
    We show mIoUs over MACs for the same architectural hyperparameters, but for different sets of $\lossweight$.
    ItNets trained by considering all outputs, i.e.\ $\lossweight=1$ for all $\numlayersindex$ (purple circles), are compared to
    \capenumfont{a, e} networks with only a single output ($\lossweight=1$ for $n=1, 6, 11, 16$ for dark blue, light blue, orange and green data points, respectively, and $\lossweight=0$ otherwise),
    \capenumfont{b, f} networks with only late outputs ($\lossweight=1$ for last 6 and 12 outputs for orange and green data points, respectively),
    \capenumfont{c, g} networks with thinned out outputs ($\lossweight=1$ for every third and fifth output for orange and green data points, respectively),
    and \capenumfont{d, h} networks with weighted outputs ($\lossweight$ is \emph{incr}eased from 1 to 16).
  }
  \label{fig:results_ablation}
\end{figure}

We study the impact of intermediate network outputs on the network performance by training the same network architecture with different sets of weight factors $\lossweight$ of the loss function (see \Cref{eq:loss}).
In summary, later network outputs benefit from the additional loss applied to earlier outputs.
This is supported by ablation studies, in which we trained networks with single, only late, or thinned out outputs.
The following results and conclusions are similar for both network types with and without weight sharing and, hence, we discuss them jointly.

The training of single outputs is worse than jointly training all outputs, except for only training the first output (see \Cref{fig:results_ablation}a and e).
However, a network with only the first output has no practical relevance due to its small network size and, consequently, low mIoU.

The intermediate activations of the early layers are optimized for the potentially conflicting tasks of providing the basis for high accuracy at early outputs and, at the same time, top accuracy at late outputs.
However, our experiments suggest that especially the optimization of the early outputs improves the performance of late outputs (see \Cref{fig:results_ablation}b and f).
Our observation that early losses do not conflict with late losses allows to keep these early losses resulting in networks with very low latency (see \Cref{fig:results_kpi}d).
In addition, keeping the first output to allow for low latency, but decreasing the density of outputs, removes unnecessary constraints and slightly improves the mIoU (see \Cref{fig:results_ablation}c and g).
The removed outputs are likely not required in applications, since the improvement of the mIoU between consecutive outputs is rather small.
Consequently, we thin out the outputs for our results shown in \Cref{fig:results_kpi}.

So far, we removed sets of outputs from the loss function, but kept the weight factors $\lossweight$ identical for all remaining outputs.
For linearly increasing the weight factors over the outputs, the mIoU jointly decreases for all outputs (compare different colors in \Cref{fig:results_ablation}d and h).
This unexpected decrease in the mIoU for late outputs may be attributed to the observed effect that earlier outputs are crucial for late performance as discussed above.

\subsection{Comparing the performance of ItNets to the literature}
\label{sec:kpis}
\begin{figure}[t]
  \centering
  \ifgcpr
\else
    \includegraphics[scale=1.0]{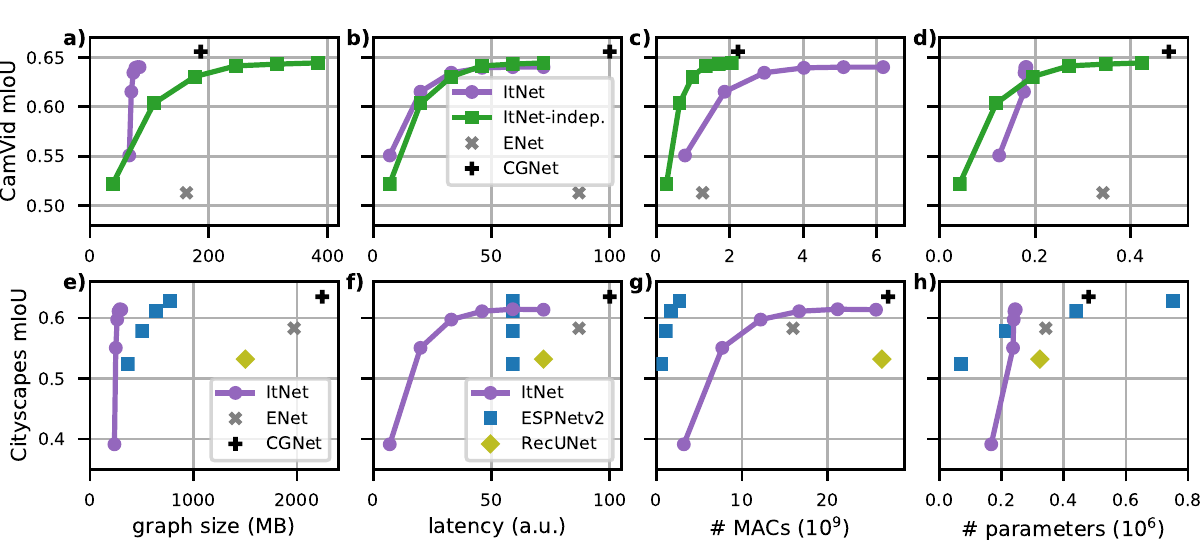}
  \fi
  \caption{
    Network performance in terms of mIoU over
    the size of the computational graph (a, e),
    the latency (b, f),
    the number of MACs (c, g),
    and the number of parameters (d, h)
    on the test set of the \camvid{} (top)
    and the validation set of the \cityscapes{} dataset (bottom).
    %
    Note that data points for ESPNetv2 without pre-training are taken from \cite{Ruiz2019_mixture}.
    %
  }
  \label{fig:results_kpi}
\end{figure}

Iterative networks require less than half the size for the computational graph compared to networks that are state-of-the-art in terms of the lowest number of MACs at comparable accuracy (compare 276MB for ItNet to 639MB for ESPNetv2 to achieve a mIoU of $61.1$ for the \cityscapes{} dataset as shown in \Cref{fig:results_kpi}e).
In addition, the multiple intermediate network outputs allow for anytime prediction and a lower latency that is further reduced by the shallow network design (\Cref{fig:results_kpi}f).
However, the number of MACs for this ItNet is approximately ten times larger than for the ESPNetv2 (16.7 compared to 1.7 billion MACs in \Cref{fig:results_kpi}g).
For the ItNet trained on \cityscapes{} (\Cref{fig:results_kpi}), exemplary network outputs are shown in \Cref{fig:intro}b (validation sample with $75$th-percentile mIoU).

By design, ItNets and ESPNetv2 have fundamentally different computational graphs and it is unclear which differences have the biggest effect on the network performance.
Since introducing a loop into the computational graphs, as done for ItNets, has the biggest effect on the size of the computational graphs,
we compare networks with and without such loops exemplarily for the \camvid{} dataset.
To exclude effects of other differences between these two types of networks,
we compare ItNets, for which the weights are shared between iterations, to ItNets with identical hyperparameters but independent weights.
Compared to ItNets with shared weights, ItNets with independent weights have substantially more free parameters that, consequently, results in better mIoUs (\Cref{fig:results_size}b).
To compensate for this effect we decrease the width of ItNets with independent weights from $\numoutchan=51$ to $\numoutchan=29$
resulting in a maximum mIoU (out of $5$ trials) of $72.7$ which is comparable to that of ItNets with shared weights ($72.8$; see data points highlighted in red in \Cref{fig:results_size}b).
Then, compared to ItNets with shared weights, ItNets with independent weights have $233\%$ the number of parameters (compare 424 to 182 thousand parameters to achieve a mIoU of approximately $64.2$ for the \camvid{} test set) and approximately three times less MACs (compare 2.0 to 6.2 billion MACs as highlighted in red in \Cref{fig:results_size}b).
The size of the computational graph is dominated by the number of unique nodes and their size.
By untying the weights between the iterations of the iterative block, the loop in the network graph has to be unrolled over its iterations, which substantially increases the number of unique nodes and, hence, the computational graph by a factor larger than $4$ (compare 83MB to 384MB in \Cref{fig:results_kpi}c).
The latency is identical for all networks, since we use the same architectural hyperparameters (\Cref{fig:results_kpi}d).

\section{Discussion}\label{discussion}
\label{sec:discussion}
In this study, we introduce a new class of network models called \emph{iterative neural networks} that have loops in their computational graphs to reduce their memory footprints and, hence, to enable their execution on massively-parallel hardware systems.
We investigated the trade-off between the size of the computational graphs and the prediction accuracies, and showed that ItNets achieve state-of-the-art performance for this trade-off (see \Cref{fig:results_kpi}e).
However, the reduction of the computational graph comes with an increase in the number of MACs (see \Cref{fig:results_kpi}g), which is expected to be smaller than the increase in throughput offered by novel, massively-parallel hardware accelerators, especially if the computational graphs fit into the in-computation memory of these hardware systems (see also \Cref{sec:intro}).
For example, \cite{graphcore_benchmark} reports a $600$-fold increase in throughput and a $16$-fold decrease in latency compared to GPUs for recurrent and convolutional networks, respectively.
In terms of energy efficiency, \cite{Esser2016_tnapps} report more than $5000$ frames per second and watt for computer vision tasks.
During the training of ItNets, we observe that additional and especially early network outputs improve the performance of late outputs, which was also observed by \cite{Zhou2019_elastic} on pre-trained networks for age estimation.
In principle, the presented methods could also be applied to different tasks, like object classification and detection, for which we expect similar observations.
However, these datasets are usually provided in a lower spatial resolution, which reduces the challenge and the need to reduce the size of these already rather small computational graphs.

Showing measurements on actual hardware is difficult, since massively systems are not easily accessible at the moment and, especially if analog circuits are involved, impose additional hardware-specific restrictions like, e.g., limited connectivity, limited precision, fixed-pattern noise and trial-to-trial variability.
The re-configuration of weights, e.g.\ the re-programming of cross-bar arrays in mixed-signal systems, is time- and energy-consuming and diminishes the benefits of massively-parallel hardware systems, namely: high throughput, low latency, and low power.
Only if the full network graph fits into the in-computation memory, independent nodes in this graph can be executed in parallel and input data can be pipelined or, in other words, processed in a streaming fashion (see also \cite{fischer2018_streaming}).
To this end, ItNets significantly reduce the footprint of their graphs by introducing loops and, hence, offer the best trade-off between model accuracy and graph size.
For the \cityscapes{} dataset, our ItNet already requires two \texttt{Nvidia V100} GPUs with 32GB memory each if the standard training procedure is not modified.
Consequently, exploring large-scale ItNets is difficult by using backpropagation on GPUs.
However, since ItNets are optimized for their execution on massively-parallel hardware systems,
this demonstrates the fundamentally different operation principles between these systems and GPUs as well as highlights the need to think beyond workloads that are tailored for GPUs.

In parallel to this study, a novel training technique was introduced by \cite{bai2020_multiscale} that significantly improves the scaling of vision networks composed of iteratively executed building blocks with weight sharing.
Instead of using backpropagation, they optimize the equilibrium point of a fix-point process described by the iterative execution of the building block.
They achieved task accuracies comparable to state-of-the-art networks and, to this end, required approximately the same number of parameters compared to the baselines (8 to 71 million parameters for \cityscapes{}).
In line with our study, this results in comparably large building blocks that are executed many times and, consequently, require significantly more MACs than the baseline networks\todominor{add details?}.
However, the training method of \cite{bai2020_multiscale} can not be applied to networks, for which the weights are allowed to be updated between iterations and, hence, the effect of weight sharing can not be studied in isolation.
In addition, their method includes the raw image (8MB in \texttt{float32} for \cityscapes{}) into the state of the iterative block, which sets a lower bound on the network size and prevents to shrink their networks to the regime of few billion MACs as shown by our method.
In contrast to \cite{bai2020_multiscale}, the iterative block of ItNets operates on compact intermediate representations with a size of only 4.8MB which is already smaller than only the raw image.

The concept of reusing convolutional building blocks has, previously, also been applied to predict semantic maps \cite{pinheiro_2014} and to generate super-resolution images \cite{Cheng_2018}.
However, instead of reusing a building block that processes all spatial scales in parallel like in the study at hand, these works reuse the same building block for different spatial scales, which does not allow for anytime predictions.

As an alternative to implementing the recurrence by convolutions as presented in this study,
\cite{Valipour2017} and \cite{Wang_2019_ICCV} connect intermediate feature maps of convolutional encoder-decoder networks by recurrent units, like LSTMs,
to predict semantic maps for the frames of videos or to consecutively improve the prediction of semantic maps for single images, respectively.
\cite{ballas_2016} apply a similar technique to encoder networks for video tasks.
However, this alternative approach has a more heterogeneous computational workload and usually results in bigger network graphs with lower accuracy (see \Cref{fig:results_kpi}e).

As an extension of this study, partly releasing the constraint of weight sharing may allow for a significant reduction in the required size of the iterative block.
To this end, a ratio between shared and free weights may be chosen close to, but smaller than one.
This will require to reconfigure the network graph on hardware, which may be costly or not possible at all depending on the hardware system.
An alternative approach to release constraints would be to replace the weights of the building block by a functions that modulates the weights over the iterations of the building block.
Another interesting open question for further studies is the root cause for the observation that early outputs improve the performance of late outputs.
This effect may be attributed to some kind of knowledge distillation and / or to the shortcut for the gradients from network output to input.

\ifgcpr
\else
\section*{Acknowledgment}
Thank you Anna Khoreva, William Beluch and Thomas Elsken for fruitful discussions and thank you Robin Hutmacher for your technical support.
This publication has received funding from the European Union's Horizon 2020 research innovation programme under grant agreement 732642 (ULPEC project).

\bibliography{bib/refs}
\bibliographystyle{abbrvnat}
\fi

\clearpage
\appendix
\renewcommand{\thefigure}{A\arabic{figure}}
\setcounter{figure}{0}
\section{Appendix}
\todominor[inline]{add comment that learning curves visually checked}
\begin{figure}
  \centering
  \includegraphics{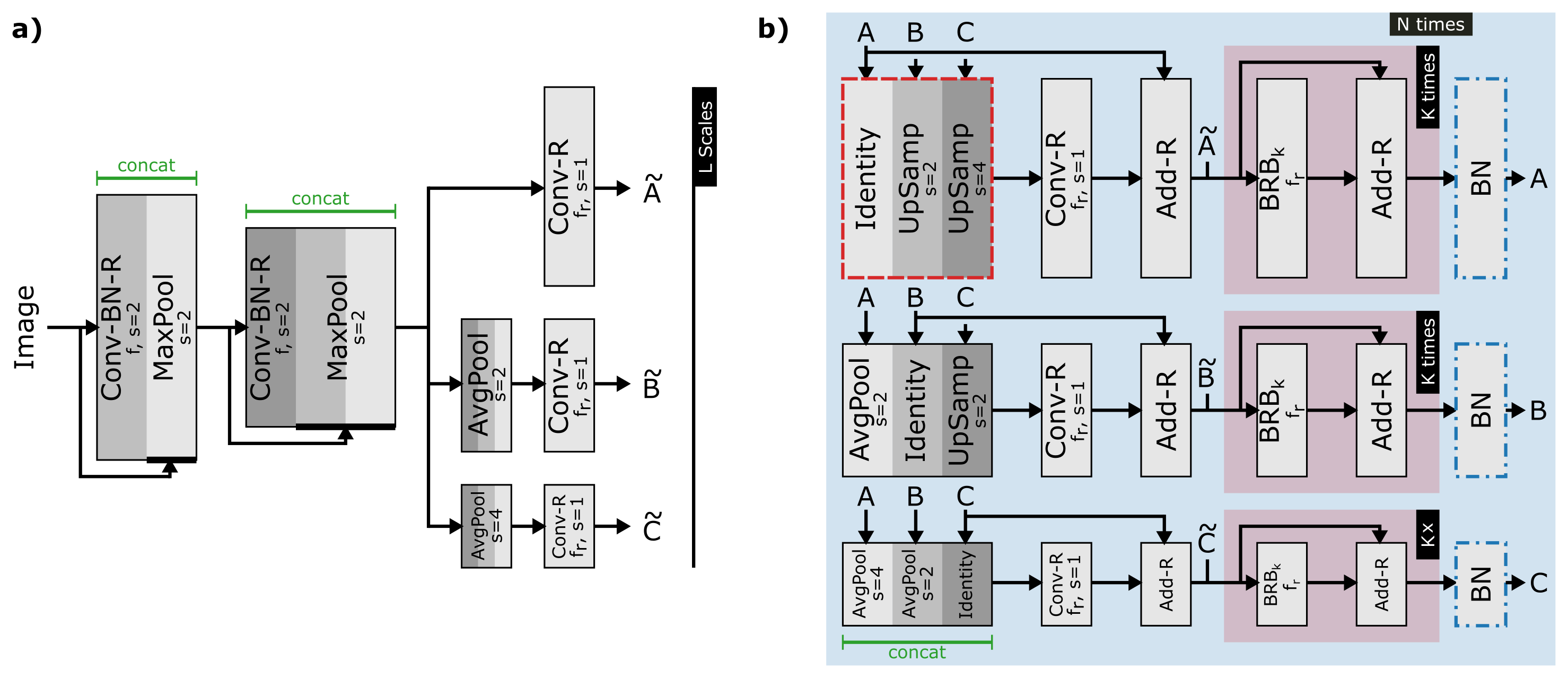}
  \caption{
    Detailed description of the \emph{data block} (a) and \emph{iterative block} (b) as shown in \Cref{fig:intro}:
    \capenumfont{a} First, images are down-sampled in two stages, each using a stride and pooling of size $s=2$ (inspired by \cite{paszke2016_enet}).
      Then, the output of the second stage is processed on $\numblocks$ different scales.
      For clarity, an example with $\numblocks=3$ is shown.
      Convolutional layers (with kernel size 3), batch-normalization layers and ReLU activation functions are denoted with \texttt{Conv}, \texttt{BN} and \texttt{R}, respectively.
    \capenumfont{b} The iterative block receives input at $\numblocks$ different scales denoted as \texttt{A}, \texttt{B} and \texttt{C}.
      In order to mix the information between the different scales (e.g., \cite{Zhao2017_pspnet}) the inputs \texttt{A}, \texttt{B} and \texttt{C} are concatenated and processed with a residual block \cite{He2016_resnet}.
      Then, an additional $\numrepeats$ \emph{bottleneck residual blocks} (BRB; with expansion factor $t=6$ \cite{Sandler2018_mobilenetv2}) are applied to obtain the output of the iterative block consisting of feature maps at again $\numblocks$ different scales.
      This output is fed back as input for the next iteration $\numlayersindex$.
      The classification block processes the concatenation of these feedback signals (highlighted in dashed red) with a $1 \times 1$ convolutional layer with $C$ output channels and, then, bilinearly up-samples (8-fold) this output to the original spatial dimensions.
      In the first iteration ($\numlayersindex=1$) of the iterative block, the mixing of the scales is skipped and the input is directly processed by the $\numrepeats$ bottleneck residual blocks (see injection of input as $\tilde{\texttt{A}}$, $\tilde{\texttt{B}}$ and $\tilde{\texttt{C}}$).
      The number of output channels are denoted by $\numoutchan$ and $\numoutchanres = \numoutchan / 2$, respectively.
      All parameters of this iterative block (blue box) are shared between iterations $\numlayersindex$, except the batch-normalization layers (dotdashed red; similar to \cite{Guo2019_CVPR}).
  }
  \label{fig:supp_methods}
\end{figure}
\begin{figure}
  \centering
  \includegraphics[scale=1.0]{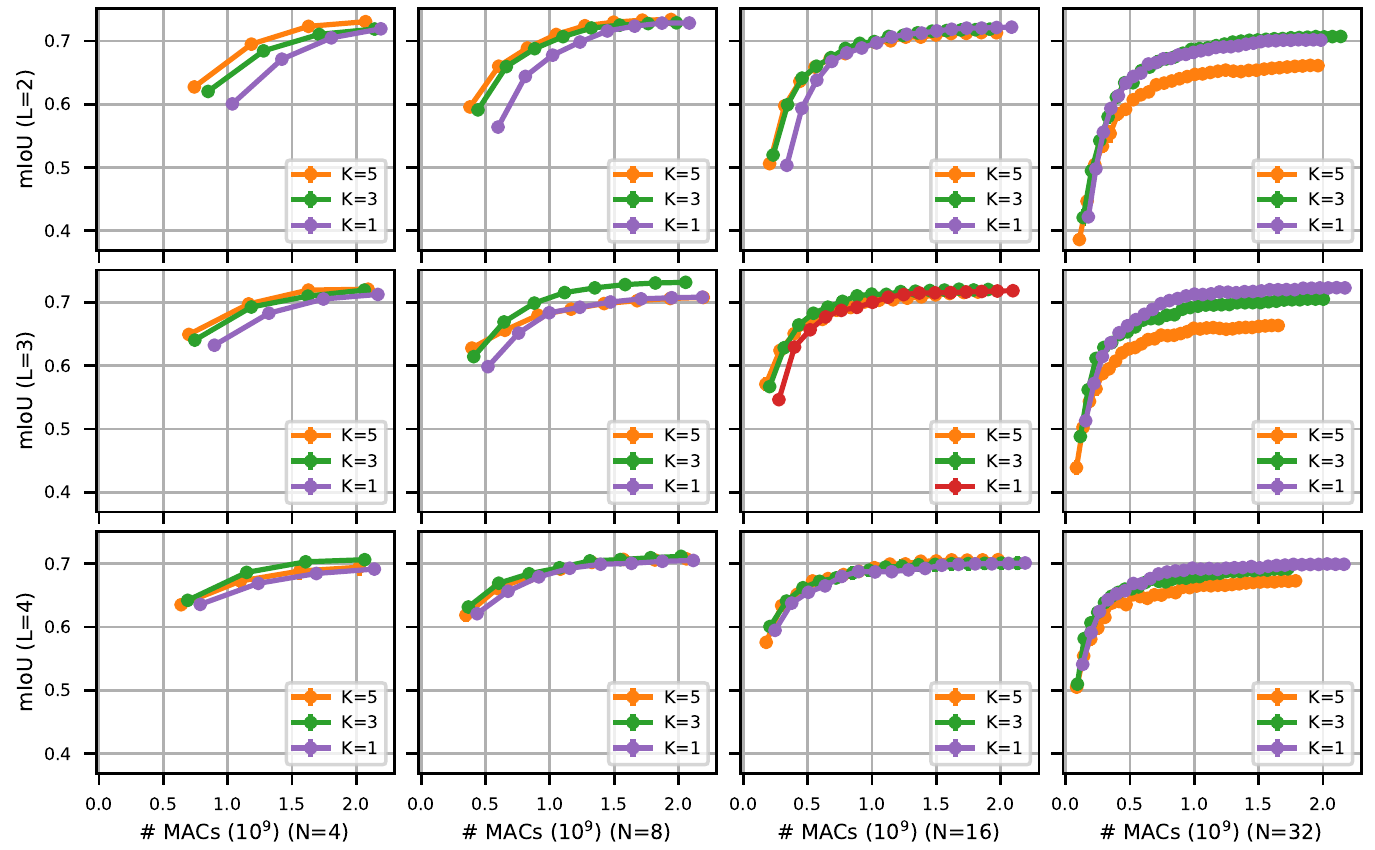}
  \caption{
    Grid search over the following hyperparameters on the validation set of the \camvid{} dataset: number of layers $\numlayers$, number of blocks $\numblocks$ and number of residual blocks $\numrepeats$.
    Parameters are not shared between iterations $\numlayersindex$ of the iterative block.
    The best set of hyperparameters (for details about the selection, see \Cref{sec:nas}) is depicted in red.
  }
  \label{fig:grid_non_shared}
\end{figure}
\begin{figure}
  \centering
  \includegraphics[scale=1.0]{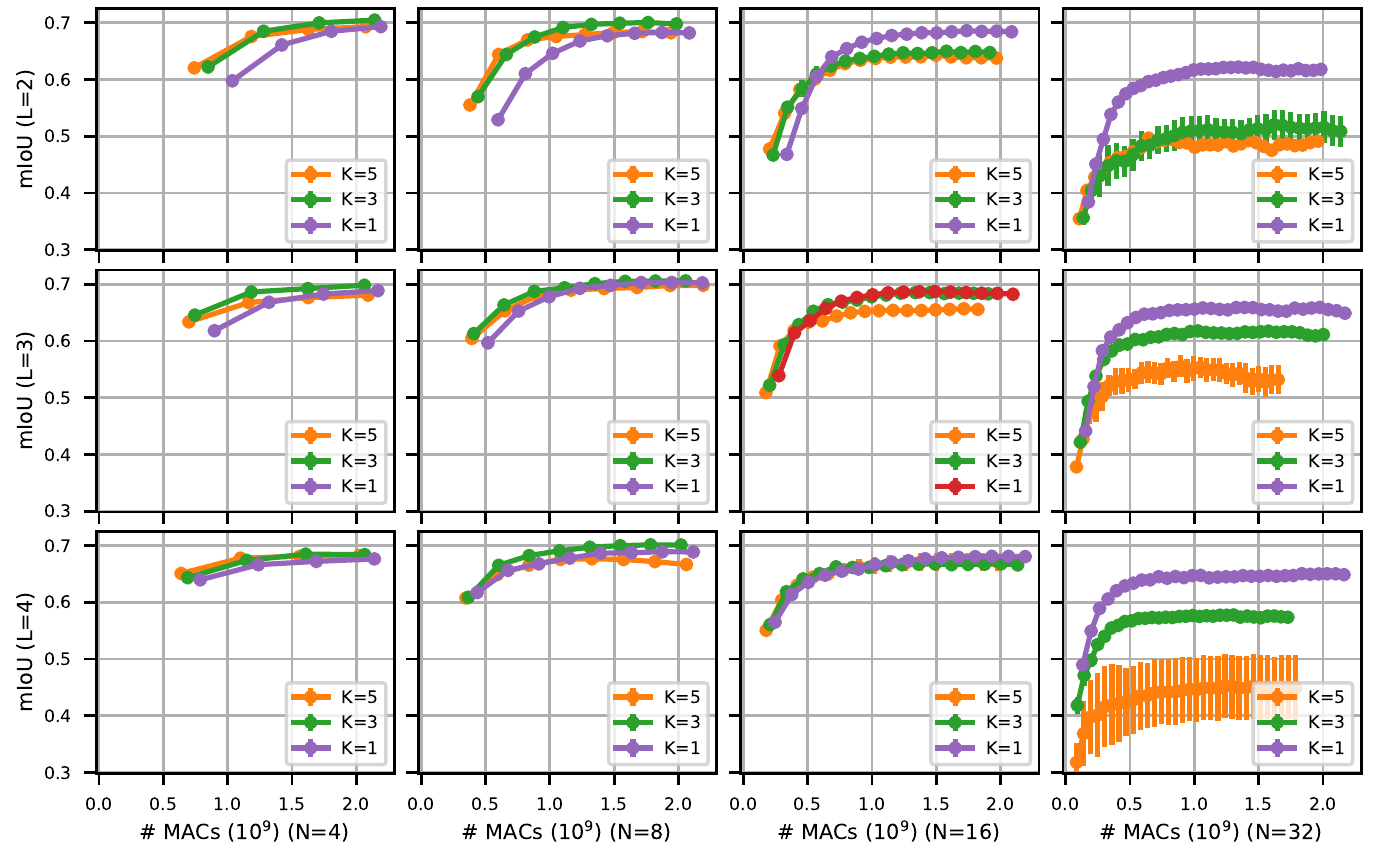}
  \caption{
    Like \Cref{fig:grid_non_shared}, but the parameters are shared between iterations $\numlayersindex$ of the iterative block.
  }
  \label{fig:grid_shared}
\end{figure}

\end{document}